\documentclass{article}

\usepackage{microtype}
\usepackage{graphicx}
\usepackage{subfigure}
\usepackage{booktabs}
\usepackage{amsmath}
\usepackage{amssymb}
\usepackage{multirow}
\usepackage{booktabs}
\usepackage{bm}
\usepackage{dsfont}
\usepackage{array}
\usepackage{times}
\usepackage{epsfig}
\usepackage{float}
\usepackage{footnote}

\usepackage[T1]{fontenc}
\usepackage{aecompl}

\usepackage{hyperref}

\usepackage[accepted]{icml2020}

\icmltitlerunning{Submission and Formatting Instructions for ICML 2020}

\begin{document}

\twocolumn[
\icmltitle{MoNet3D: Towards Accurate Monocular 3D Object Localization in Real Time}



\icmlsetsymbol{equal}{*}

\begin{icmlauthorlist}
\icmlauthor{Xichuan Zhou}{to}
\icmlauthor{Yicong Peng}{to}
\icmlauthor{Chunqiao Long}{to}
\icmlauthor{Fengbo Ren}{ed}
\icmlauthor{Cong Shi}{to}
\end{icmlauthorlist}

\icmlaffiliation{to}{Key Laboratory of Dependable Service Computing in Cyber Physical Society Ministry of Education, College of Microelectronics and Communication Engineering, Chongqing University, Chongqing, China 400044.}
\icmlaffiliation{ed}{Arizona State University, Tempe, Arizona, United States}

\icmlcorrespondingauthor{Xichuan Zhou}{zxc@cqu.edu.cn}

\icmlkeywords{Autonomous Driving, 3D Object Detection, Neural Network}

\vskip 0.3in
]



\printAffiliationsAndNotice{}  

\begin{abstract}
Monocular multi-object detection and localization in 3D space has been proven to be a challenging task. The MoNet3D algorithm is a novel and effective framework that can predict the 3D position of each object in a monocular image and draw a 3D bounding box for each object. The MoNet3D method incorporates prior knowledge of the spatial geometric correlation of neighbouring objects into the deep neural network training process to improve the accuracy of 3D object localization. Experiments on the KITTI dataset show that the accuracy for predicting the depth and horizontal coordinates of objects in 3D space can reach 96.25\% and 94.74\%, respectively. Moreover, the method can realize the real-time image processing at 27.85 FPS, showing promising potential for embedded advanced driving-assistance system applications. Our code is publicly available at \url{https://github.com/CQUlearningsystemgroup/YicongPeng}.
\end{abstract}

\section{Introduction}
In recent years, computer vision-based automated driving-assistance technology has made great progress. The rapid development of deep learning-based methods has enabled researchers and engineers to develop accurate and cost-effective advanced driving-assistance systems (ADASs), for which object detection and localization is one of the key functions. Various methods based on convolutional neural networks (CNNs) have been proposed for 2D object detection from monocular video images\cite{girshick2014rich,redmon2016you,liu2016ssd}. However, despite its great advantages in terms of efficiency and cost, 3D object detection based on monocular vision is still greatly challenging.

Compared with solutions such as LiDAR and stereo vision, the accuracy of the monocular method is far from sufficient for ADAS applications. For example, when using the KITTI 3D object detection benchmark to detect the category of cars, the average accuracy of the state-of-the-art monocular vision algorithm is 63.02\% lower than that of LiDAR-based algorithms\cite{bao2019monofenet,shi2020pv}.

Using monocular and single frames of RGB images for 3D object localization and detection can reduce the hardware cost of ADAS applications, but it also brings great technical challenges. First, the images captured by monocular images lack depth-of-field information, and in principle, it is difficult to achieve 3D object localization. Second, different degrees of vehicle occlusion, lack of image information, inelastic distortion caused by rotating the target object, and distortion caused by lens imaging all make monocular 3D object localization more challenging. To meet these challenges, this paper establishes a neural network called MoNet3D by introducing the geometric relationship of neighbouring objects in 3D space to improve the accuracy of 3D object detection and localization.

Specifically, to cope with the 3D localization problem with severely insufficient constraints, some researchers have recently attempted to use prior knowledge to optimize deep learning methods. For example, 3D-Deepbox uses prior knowledge that the predicted 3D bounding box should closely fit the 2D bounding box\cite{Mousavian}. Mono3D$\_$PLiDAR relaxed this constraint, assuming that the 2D projection of a 3D object is globally consistent with the bounding box of the 2D object\cite{weng2019monocular}. These studies show that the geometric relationship between the 2D and 3D bounding boxes associated with detected objects can help to achieve 3D object localization, but their assumption of global consistency might not be met in the face of various types of noise, such as inelastic distortion, and their experimental results show that the research on monocular 3D positioning is still in an early stage.

To address this challenge, we relax the assumption of \emph{global} geometric  consistency. Instead, MoNet3D attempts to incorporate prior knowledge of the \emph{local} geometric consistency. Intuitively, the proposed method is based on the observation that, given a pair of objects with similar depths, if they are close to each other in the image, they should also be close to each other in actual 3D space. Therefore, the local geometric relations should be helpful for guiding the prediction of 3D object localization. From a methodological point of view, MoNet3D is an end-to-end deep neural network that consists of three stages. The first stage extracts multi-layer features from the image for object detection and localization. The second stage detects 2D objects from monocular images, and the features of the 2D objects are sent to the third stage for 3D object localization. The local consistency of neighbouring objects is formalized as a regularization term to constrain the prediction of 3D localization. By incorporating prior knowledge of local consistency, MoNet3D can improve the accuracy and convergence speed of the deep network training process.

In summary, the main advantages and contributions of MoNet3D are four-fold.
 \begin{itemize}
\item \emph{Accurate 3D object localization}: By incorporating prior knowledge of the 3D local consistency, MoNet3D can achieve 95.50\% accuracy on average for 3D object localization.
\item \emph{More accurate 3D object detection}: MoNet3D achieves 3D object detection accuracy of 72.56\% in the KITTI dataset (IoU=0.3), which is competitive with state-of-the-art methods.
\item \emph{High efficiency}: MoNet3D can process video images at a speed of 27.85 frames per second for 3D object localization and detection, which makes it promising method for embedded ADAS applications.
\item \emph{Open source}: Part of the data and code of MoNet3D will be publicly available on the GitHub website when the paper is published.
 \end{itemize}

\section{Related Work}

\subsection{3D Object Detection from LiDAR}
Most existing studies of 3D object detection are based on LiDAR sensors \cite{li2016vehicle}. More recently, with the development of deep learning methods, Qi proposed using a deep neural network for 3D object detection with point cloud data \cite{qi2017pointnet,qi2017pointnet++,qi2018frustum}. Later, Zhou divided the point cloud into 3D voxels and converted the set of points in each voxel into a single feature representation through the voxel-feature coding layer\cite{zhou2018voxelnet}. Chen proposed the MV3D method, which combines vision and LiDAR point cloud information.\cite{chen2017multi}. Although these algorithms achieve state-of-the-art results for 3D object detection, they are rarely applied for ADAS applications due to economic reasons.

\subsection{3D Object Detection for a Single Monocular Image}
Instead of installing expensive LiDAR-based systems for 3D object detection, many level-three autonomous cars attempt to use computer vision-based approaches for 3D object detection due to their economic advantages. Very recently, Chen proposed applying deep learning in 3D object detection when using a single camera \cite{chen2016monocular}. Since then, research on monocular-based 3D object detection has attracted increasing attention\cite{fang20193d,zhuo20183d,crivellaro2017robust}. For example, Roddick proposed OFT-Net, which maps image-based features onto an orthogonal 3D space for 3D object detection \cite{roddick2018orthographic}; Liu proposed measuring the degree of visual fit between the projected 3D region proposal and the 2D object on the image \cite{liu2019deep}; Simonelli proposed using the regression loss to make the training process more stable\cite{simonelli2019disentangling}; Li improved the prediction accuracy of the 3D box method by using the fused features of visible surfaces\cite{li2019gs3d}; Qin used both deep and shallow features extracted by a convolutional neural network to improve the prediction accuracy of the centre point\cite{qin2019monogrnet}. These studies of monocular 3D object detection are very inspiring. However, thus far, the results are still below the expectation of industrious application, and the state-of-the-art accuracy on the KITTI dataset is generally less than 50\% for the category of cars.
\begin{figure}[t]
\vskip 0.2in
\begin{center}
   \includegraphics[width=0.8 \linewidth]{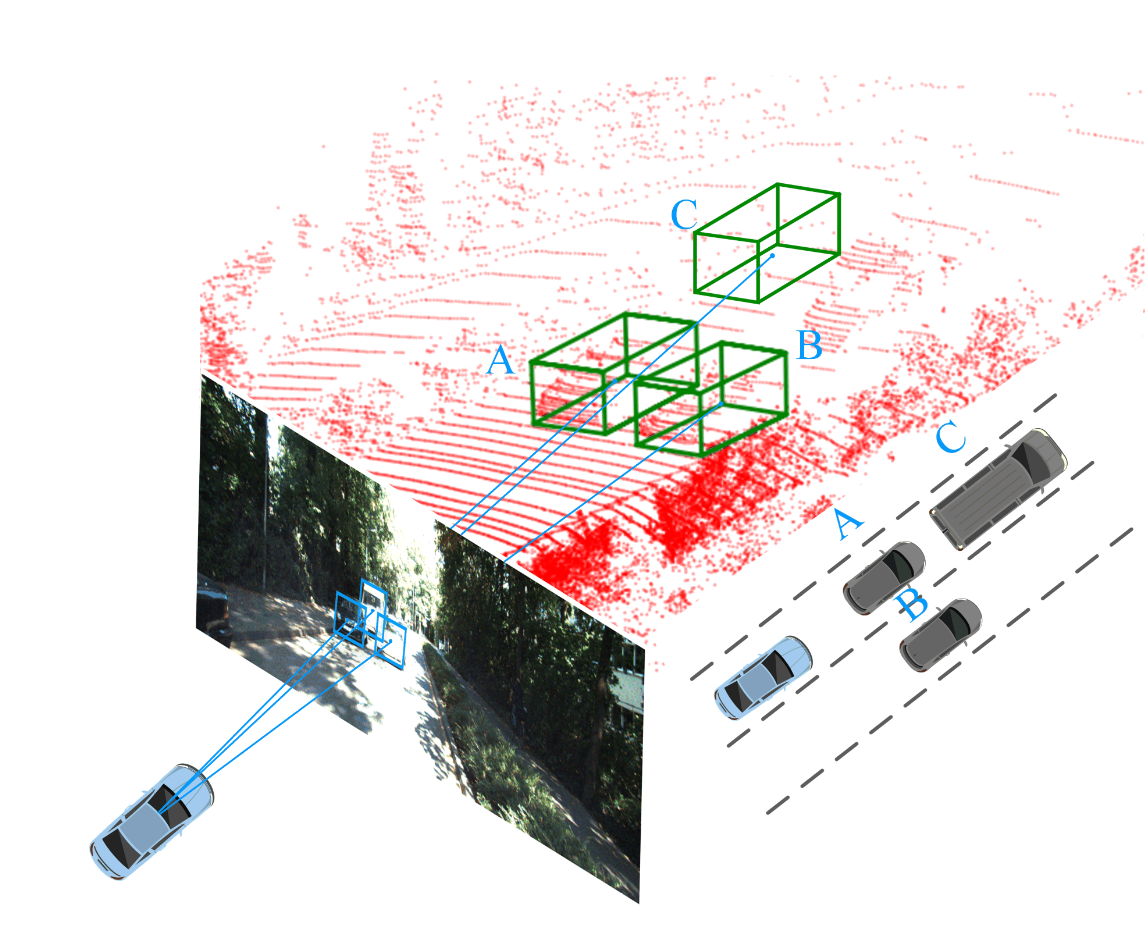}
\end{center}
\caption{An example applying MoNet3D for 3D object detection using a single RGB image. MoNet3D incorporates the horizontal neighbouring relation between cars A and C in the image, which is important for same-lane determination, to constrain the estimation of 3D localization.}
\vskip -0.2in
\label{figure1}
\end{figure}
\section{Method}
\begin{figure*}[t]
\vskip 0.2in
\begin{center}
   \includegraphics[width=1\linewidth]{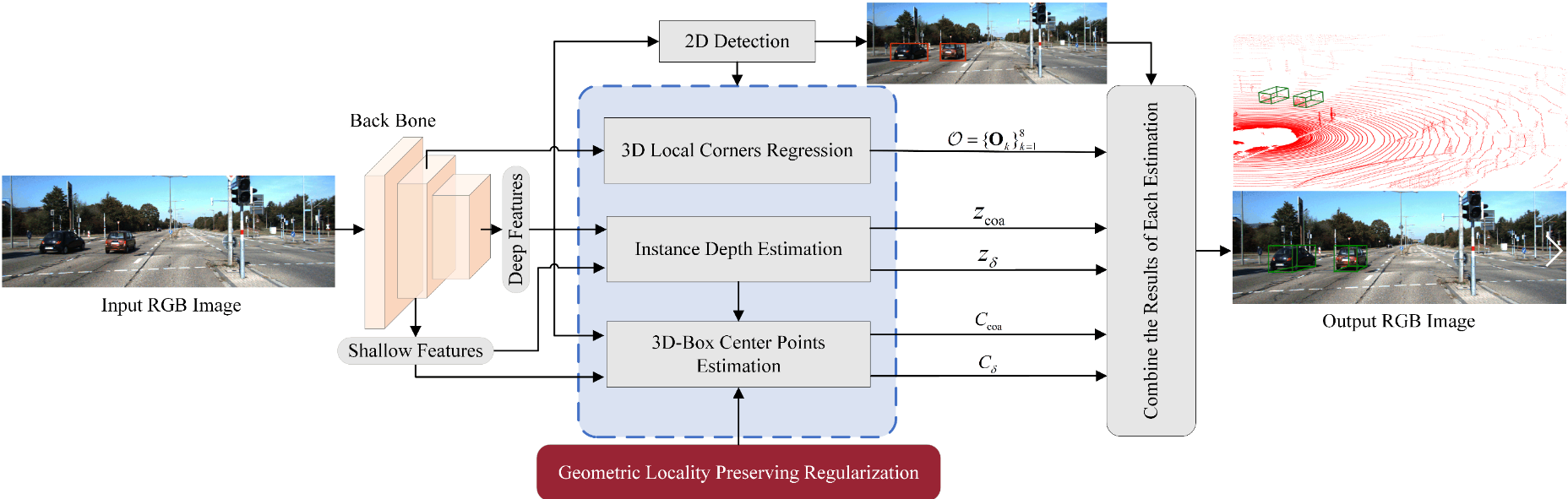}
\end{center}
\caption{ MoNet3D extracts the features from monocular RGB images for 3D object localization. It consists of four modules, including the 2D object detection module, instance depth estimation module, 3D box centre point estimation module, and 3D box corners regression module. Different from previous methods, MoNet3D uses prior knowledge of the geometric locality consistency as a regularization term to constrain the prediction of the centre point of the 3D box.}
\vskip -0.2in
\label{figure2}
\end{figure*}
To improve the accuracy of existing monocular-based 3D object detection, we propose the MoNet3D method to use the geometric correlation between neighbouring objects on the image for 3D object localization. Figure \ref{figure1} briefly illustrates our method. It can be seen for the three cars A, B and C in front of the camera car, that compared to A and B, A and C are closer on the image, which indicates that cars A and C may be closer in the real (3D) world. Based on the observation, we hope to use the horizontal distance relationship of the objects in the picture to constrain the distance of the neighbouring objects in the 3D space, so as to optimize the weight parameters in the training process of the neural network, and then improve the accuracy of 3D object localization. In practice, MoNet3D method may improve the accuracy of lane judgment, which is important for automatic driving application.

\subsection{Problem Definition}

The MoNet3D method uses a single frame of an RGB image for 3D object detection and localization. Technically, MoNet3D returns the category information, 3D position and size and of the objects of interest in the image in the form of 3D boxes. The 3D box of any object is represented by the 3D centre point $\mathrm{C}_{\mathrm{3d}} = (u^{(\mathrm{3d})}, v^{(\mathrm{3d})}, z^{(\mathrm{3d})})$ and the coordinates of the 8 vertices of the 3D box frame ${\mathbf{\mathcal{O}}} = \{\mathbf{O}_k\}_{k=1}^8$.

\subsection{Overall Network Structure}

As shown in Figure \ref{figure2}, the MoNet3D framework first use VGG-16 without the fully connected layer to extract features from a single frame of an RGB image. Similar to \cite{qin2019monogrnet}, we combine the shallow features and deep features for further object detection. The MoNet3D framework divides feature processing into four modules:
\begin{itemize}
\item The 2D detection module outputs 2D box and object recognition results based on image features and applies this information to subsequent 3D detection.
\item The instance-level depth estimation module estimates the depth information of each object and uses it for subsequent 3D box centre point estimation.
\item The 3D box centre point estimation module combines the predicted depth information and 2D box information to estimate the centre point coordinates of each 3D box. MoNet3D incorporates prior knowledge of the geometric locality as regularization for training this module.
\item The 3D local corner regression module combines the 2D recognition results and image features to regress the coordinate information of the 8 points of the 3D box frame.
\end{itemize}

It is particularly worth noting that the main challenge of this paper is the estimation of the 3D box centre point, especially the accuracy of the horizontal offset estimation. Its error determines the error in the lane determination, which plays an important role in the control and safety of autonomous driving. To improve the accuracy of the 3D box centre point estimation, MoNet3D adopts the geometric locality preserving regularization method, which is described in detail below.

\subsection{Geometric-Locality-Preserving Regularization}

To improve the 3D localization accuracy, mathematically, we formalize the assumption of geometric locality consistency as a regularization term. Suppose there are $\mathrm{M}$ objects in the training set. The matrix $\mathrm{\mathbf{S}}=\{s_{ij}\}$ defines an $\mathrm{M}\times \mathrm{M}$ similarity matrix as follows:

\begin{equation}
\label{E1}
s_{ij}\!\!=\!\exp\!\left[\!-(u_i^{(\mathrm{2d})}\!\!-\!u_j^{(\mathrm{2d})})^2\right]\!\big/\! \exp\!\left[(z_i^{(\mathrm{3d})}\!\!-\!z_j^{(\mathrm{3d})})^2\!/\!\lambda\right]
\end{equation}
where $u_i^{(\mathrm{2d})}$ and $u_j^{(\mathrm{2d})}$ are the horizontal offsets of object $i$ and object $j$, respectively, in the 2D image and $z_i$ is the ground-truth depth of object $i$.
MoNet3D assumes that when object $i$ and object $j$ have similar 3D depths, these two objects will have larger similarity $s_{ij}$ if their 2D bounding boxes have smaller horizontal offsets. Otherwise, if these two objects have a large 3D depth difference or their horizontal offset in the image is large, their geometric similarity $s_{ij}$ should be small.

To preserve the geometric similarity for predicting 3D localization, MoNet3D applies the similarity relationship defined in Equation \ref{E1} to the fully connected layer of the neural network and optimizes the 3D box centre point estimation (Figure \ref{figure2}).
Suppose the output of object $i$ in this layer is $\mathbf{y}_i = \mathrm{\mathbf{W}}\mathbf{x}_i + \mathbf{b}$, where $\mathbf{y}_i = (u_i^{(\mathrm{3d})},z_i^{(\mathrm{3d})})$, $\mathbf{x}_i$ represents the input of the fully connected layer, $\mathrm{\mathbf{W}}$ is the connection weight, and $\mathbf{b}$ is the deviation vector. Assuming that the training object $i$ and another object $j$ have large similarity values, MoNet3D attempts to optimize the connection weight $\mathrm{\mathbf{W}}$ so that objects $i$ and $j$ are close to each other in 3D space. Technically, MoNet3D minimizes the following regularization term $R (\mathrm{\mathbf{W}})$ as
\begin{equation}
\label{E2}
\arg \min_\mathbf{W} \frac{\beta}{2}\sum_{ij}\left\|\mathrm{\mathbf{W}}\mathbf{x}_i-\mathrm{\mathbf{W}}\mathbf{x}_j\right\|_2^2s_{ij}
\end{equation}
Intuitively speaking, according to the above equation, if the $i$ and $j$ object pairs are nearby with larger $s_{ij}$ values, then $s_{ij}$ would help to reduce the distance between $\mathrm{\mathbf{W}}\mathbf{x}_i$ and $\mathrm{\mathbf{W}}\mathbf{x}_j$ in the minimization process so that the similarity of object pairs in 2D space can be maintained in 3D space. For more efficient computation,
the regularization term $R(\mathrm{\mathbf{W}})$ can be equivalently written as
\begin{align}\label{E3}
&\!R\big(\mathrm{\mathbf{W}}\big)\!=\!\beta\sum_{ij}\!\mathrm{tr}\!\left[\mathrm{\mathbf{W}} \left(\mathbf{x}_i\!-\!\mathbf{x}_j\right)\!\!\left(\mathbf{x}_i\!-\!\mathbf{x}_j\right)^\mathrm{T}\!\mathrm{\mathbf{W}}\right]\!s_{ij}&
\notag \\
&=\beta \mathrm{tr}\left[ \mathrm{\mathbf{W}}^{\mathrm{T}} \mathbf{X} \mathbf{D} \mathbf{X}^{\mathrm{T}} \mathrm{\mathbf{W}} \!-\! \mathrm{\mathbf{W}}^{\mathrm{T}} \mathbf{X} \mathbf{S} \mathbf{X}^{\mathrm{T}} \mathrm{\mathbf{W}} \right]&
\notag \\
&=\beta \mathrm{tr}\left[ \mathrm{\mathbf{W}}^{\mathrm{T}} \mathbf{X} \mathbf{P} \mathbf{X}^{\mathrm{T}} \mathrm{\mathbf{W}} \right]&
\end{align}
where $\mathbf{X}=\left[ \mathbf{x}_1,\mathbf{x}_2, \dots, \mathbf{x}_{\mathrm{M}}\right]$ represents the matrix form of the input vectors of the fully connected layer. $\mathbf{D}$ is the diagonal matrix, where the element on the diagonal is $d_{ii}$: $d_{ii}=\sum_js_{ij}$, $\mathrm{\mathbf{S}}=\{s_{ij}\}$, $\mathbf{P}=\mathbf{D}-\mathbf{S}$. By applying geometric-locality-preserving regularization, MoNet3D can more accurately predict the 3D box centre point associated with each object.

\subsection{Loss Functions}
In this section, we briefly summarize the loss function of each of the four modules in the MoNet3D neural network.

\subsubsection{2D Estimation}
The MoNet3D method first estimates 2D objects in the image after feature extraction and provides region proposals for subsequent 3D object detection and localization. The 2D estimation module is a basic module that predicts and categorizes regions of interest. Here, we use fast regression from YOLO as the main estimation part and add RoIAlign to 2D estimation to improve the accuracy \cite{redmon2016you,he2017mask}. By dividing the original image into $32 \times 32$ grids (we use $\mathrm{g}$ to indicate a specific grid), we let each grid predict two 2D bounding boxes, $b_{\mathrm{2d}}^\mathrm{g} = (u^{(\mathrm{2d})}, v^{(\mathrm{2d})}, d, h)$ and the confidence $\rm{Pr}_{\rm{obj}} $, where $u^{(\mathrm{2d})}, v^{(\mathrm{2d})}, d, h$ are the coordinates of the centre point of the 2D box and the length and width of the 2D box for each cell grid $\mathrm{g}$. The final 2D box is then predicted by NMS and RoIAlign. The loss function for 2D estimation can be expressed as

\begin{align}
L_{\mathrm{2d}}&=L_{\mathrm{conf}}+\alpha L_{\mathrm{b2d}} &
\end{align}
where $L_{\mathrm{conf}} = \mathcal{E}_{\mathrm{g}}\left[ \mathcal{S}\, (\widehat{\mathrm{Pr}}_{\mathrm{obj}}^\mathrm{g}), \mathrm{Pr}_{\mathrm{obj}}^\mathrm{g}\right]$, $L_{\mathrm{b2d}}=\sum_{\mathrm{g}} \mathds{l}_{\mathrm{g}}^{\mathrm{obj}} \cdot \mathcal{L}_1 \left(  \widehat{b}_{\mathrm{2d}}^\mathrm{g}, b_{\mathrm{2d}}^\mathrm{g}, \right)$, $\rm{Pr}_{\mathrm{obj}}$ refers to the confidence of the ground truth, $\widehat{\mathrm{Pr}}_{\mathrm{obj}}^\mathrm{g}$ refers to the confidence of the predictions, $\mathcal{S}(\cdot)$ is expressed as the softmax function, $\mathcal{E}(\cdot)$ is the cross entropy, $\mathcal{L}_1(\cdot)$ is the L1 distance loss function, $\alpha$ is the balance coefficient, and $\mathds{l}_{\rm{g}}^{\mathrm{obj}}$ indicates whether there is an object in cell $\rm{g}$ (1 if there is, 0 if there is not).

\subsubsection{Instance Depth Estimation}
We let $z^{\mathrm{g}}$ denote the object depth in an arbitrary grid $\mathrm{g}$. Similar to MonoGRNet, MoNet3D combines deep and shallow features to improve the accuracy of the depth estimation network \cite{qin2019monogrnet}. MoNet3D first predicts the rough depth $z_{\mathrm{coa}}^{\mathrm{g}}$ from the deep features and then uses shallow features for fine-tuning. The final instance-level depth can be estimated as $z^{\mathrm{g}} = {z}_{\mathrm{coa}}^{\mathrm{g}} + z_{\delta}^{\mathrm{g}}$, where $z_{\delta}^{\mathrm{g}}$ is predicted by the shallow features. The loss function for estimating the depth is formalized as

\begin{align}
L_{\mathrm{z}}&= \gamma L_{\mathrm{zcoa}} + L_{\mathrm{\delta z}}
\end{align}
where $L_{\mathrm{zcoa}}=\sum_\mathrm{g} \mathds{l}_\mathrm{g}^{\mathrm{obj}} \cdot \mathcal{L}_1 \left(\widehat{z}_{\mathrm{coa}}^\mathrm{g}, z^\mathrm{g} \right)$, $L_{\mathrm{\delta z}}=\sum_\mathrm{g} \mathds{l}_\mathrm{g}^{\mathrm{obj}} \cdot \mathcal{L}_1 \left( \widehat{z}_{\mathrm{coa}}^\mathrm{g}+\widehat{z}_{\delta}^\mathrm{g}, z^\mathrm{g}  \right)$, $z^{\mathrm{g}}$ is the depth information of the ground truth, $\widehat{Z}_{coa}^{\mathrm{g}}$ is the depth information of the prediction from the deep features, $\widehat{Z}_{\delta}^{\mathrm{g}}$ refers to the object depth information of the shallow feature prediction, and $\gamma$ refers to the balance coefficient.
\begin{figure*}[!htb]
\vskip 0.2in
\centering
\begin{minipage}{5.70cm}
\centerline{High-Speed Roads}
   \includegraphics[width=1\linewidth]{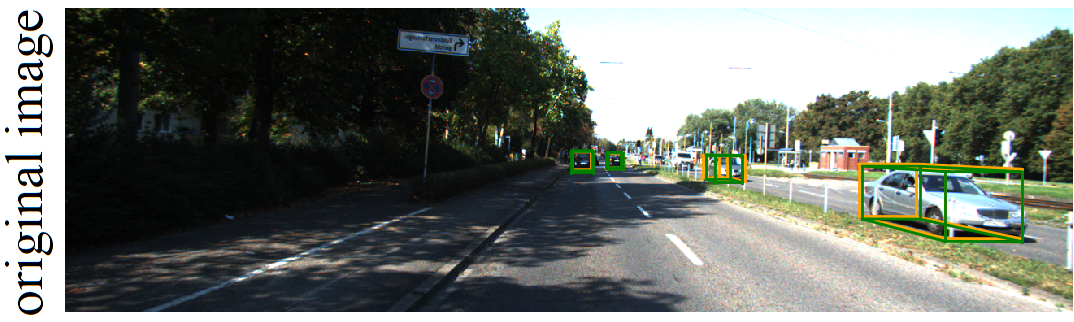}
\end{minipage}
\begin{minipage}{5.60cm}
\centerline{Town Roads}
   \includegraphics[width=1\linewidth]{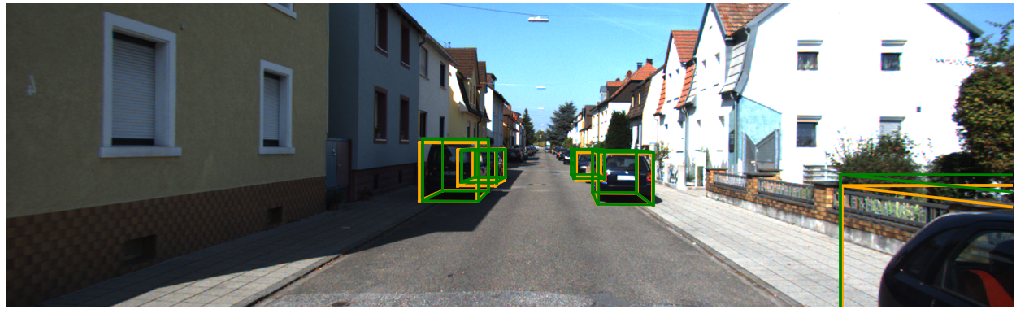}
\end{minipage}
\begin{minipage}{5.60cm}
\centerline{Neighbourhood Roads}
   \includegraphics[width=1\linewidth]{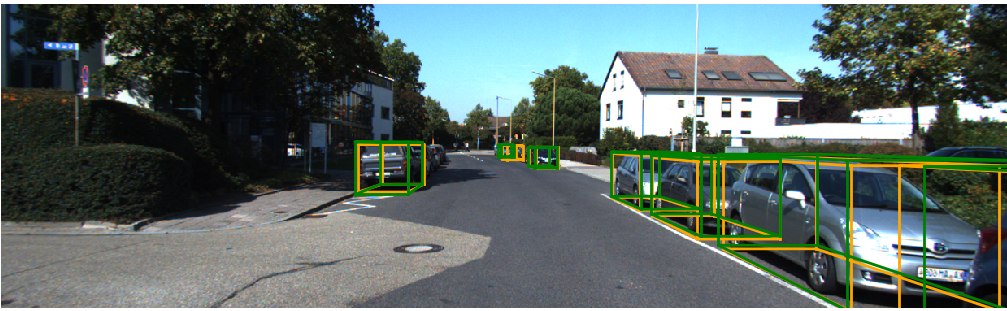}
\end{minipage}

\begin{minipage}{5.70cm}
   \includegraphics[width=1\linewidth]{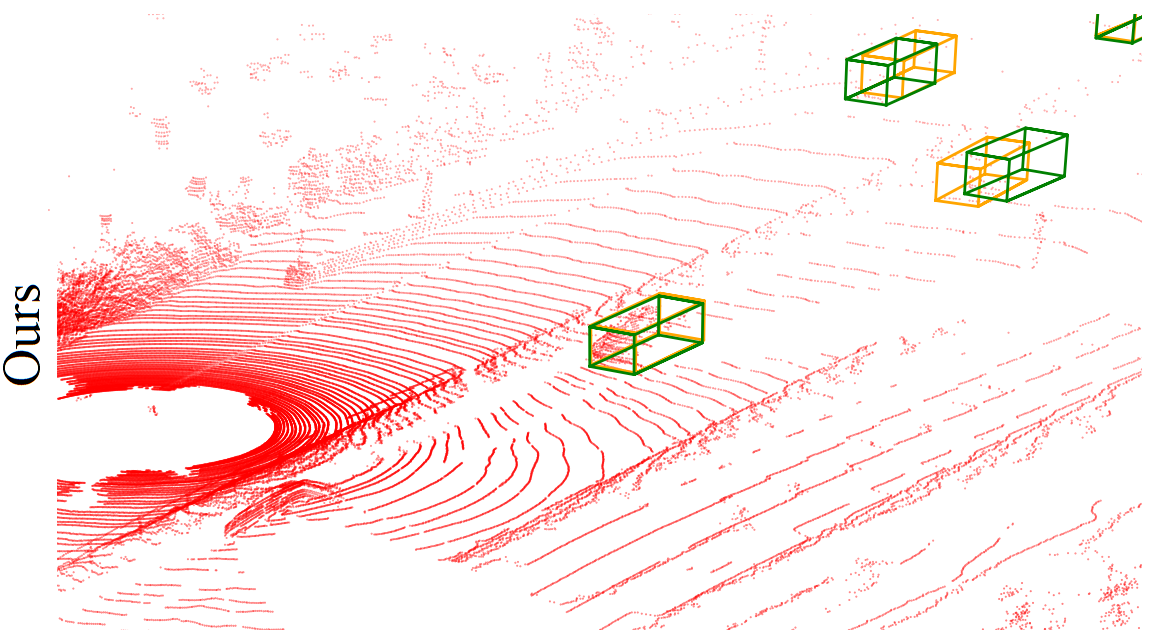}
\end{minipage}
\begin{minipage}{5.60cm}
   \includegraphics[width=1\linewidth]{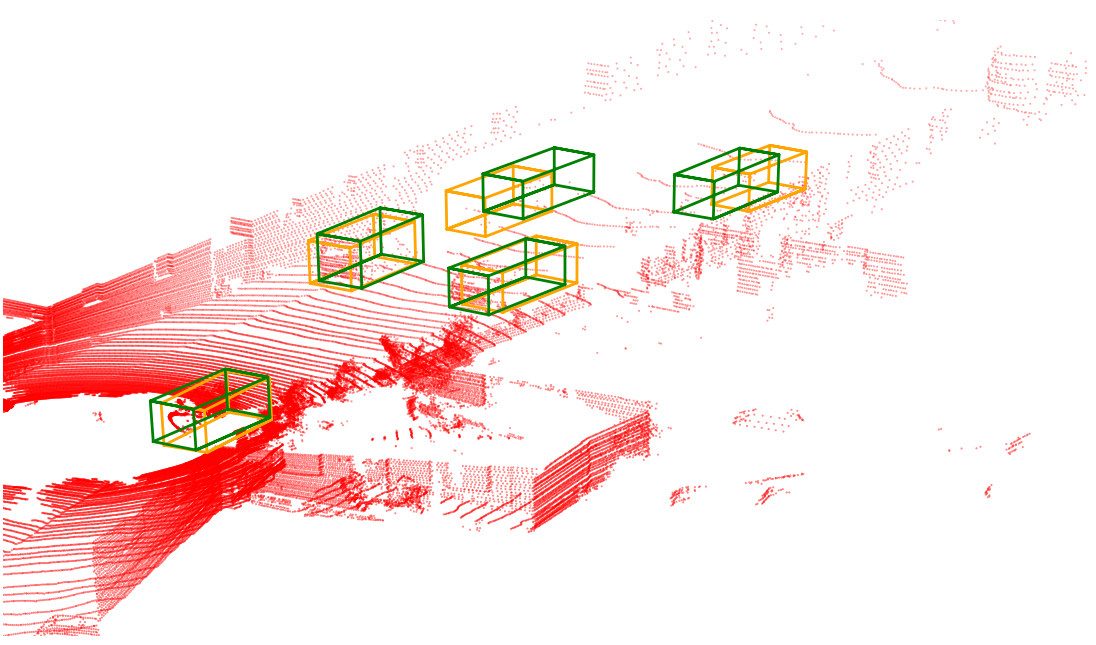}
\end{minipage}
\begin{minipage}{5.60cm}
   \includegraphics[width=1\linewidth]{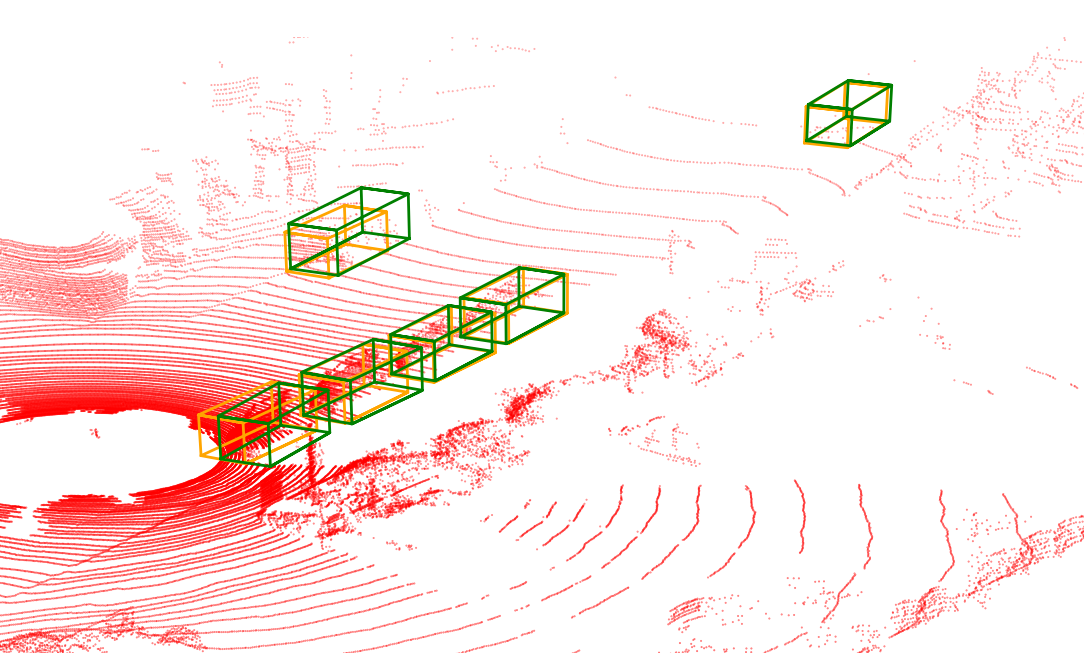}
\end{minipage}

\begin{minipage}{5.70cm}
   \includegraphics[width=1\linewidth]{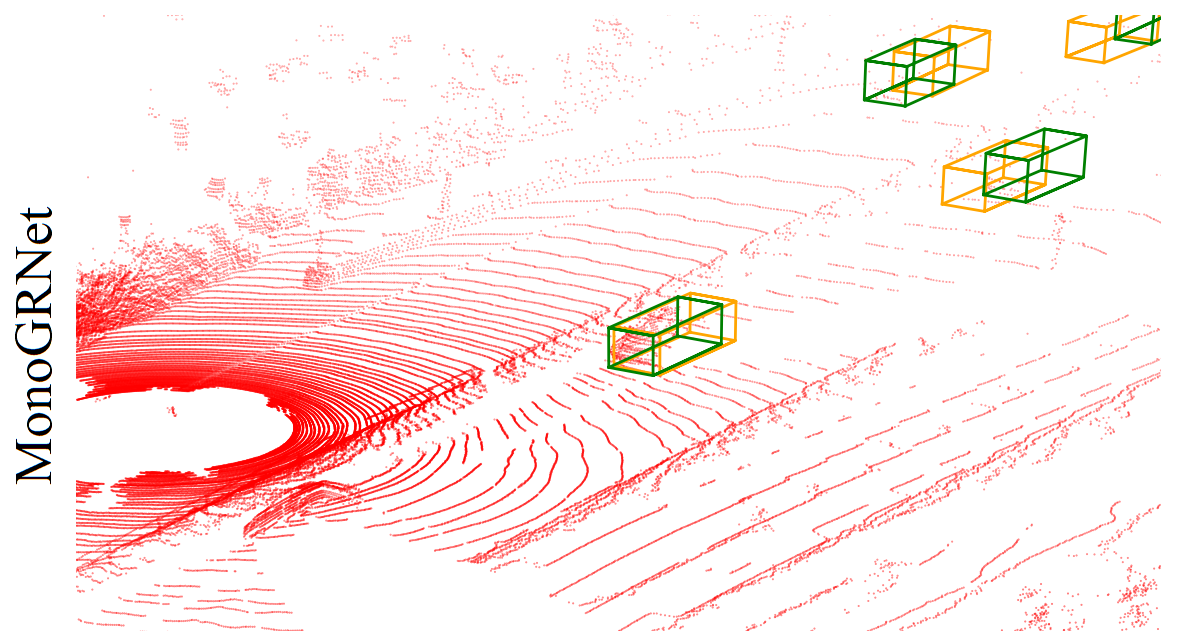}
\end{minipage}
\begin{minipage}{5.60cm}
   \includegraphics[width=1\linewidth]{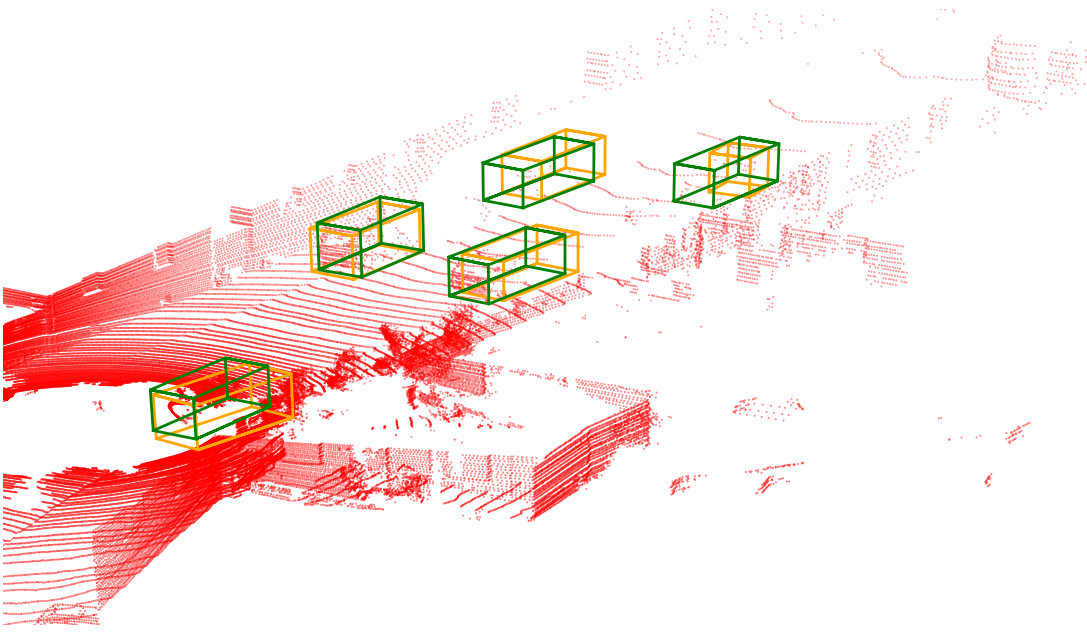}
\end{minipage}
\begin{minipage}{5.60cm}
   \includegraphics[width=1\linewidth]{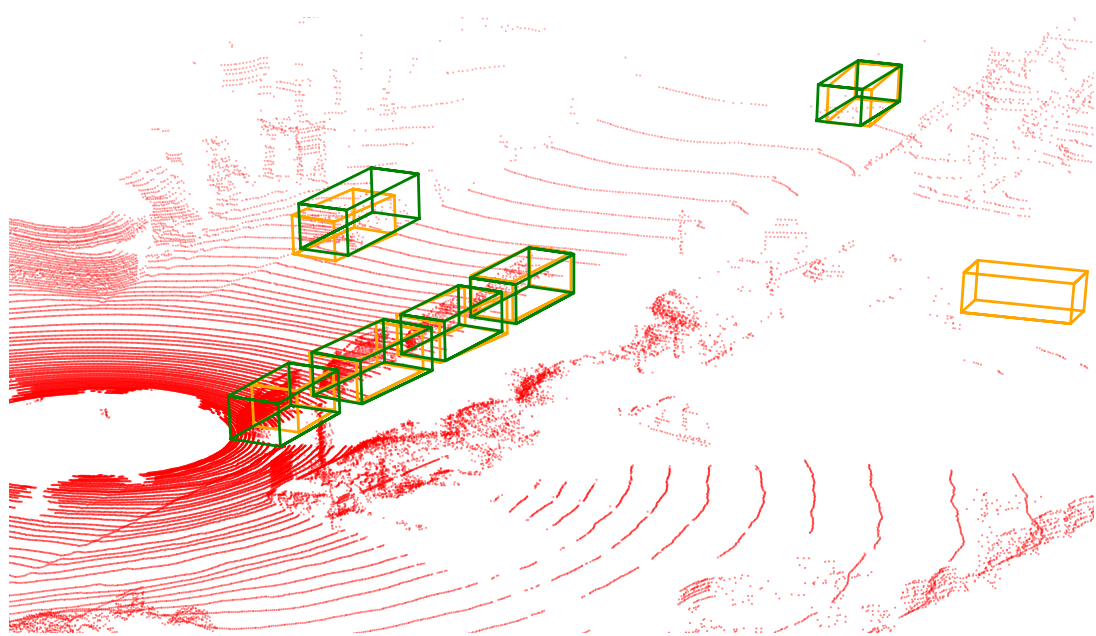}
\end{minipage}

\begin{minipage}{5.70cm}
   \includegraphics[width=1\linewidth]{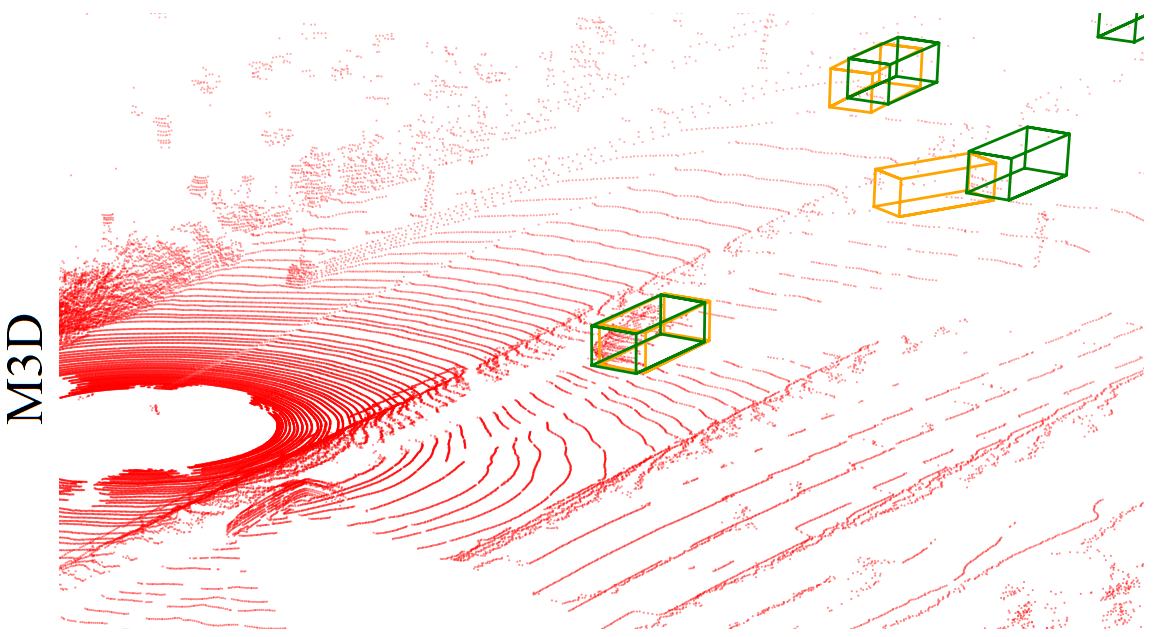}
\end{minipage}
\begin{minipage}{5.60cm}
   \includegraphics[width=1\linewidth]{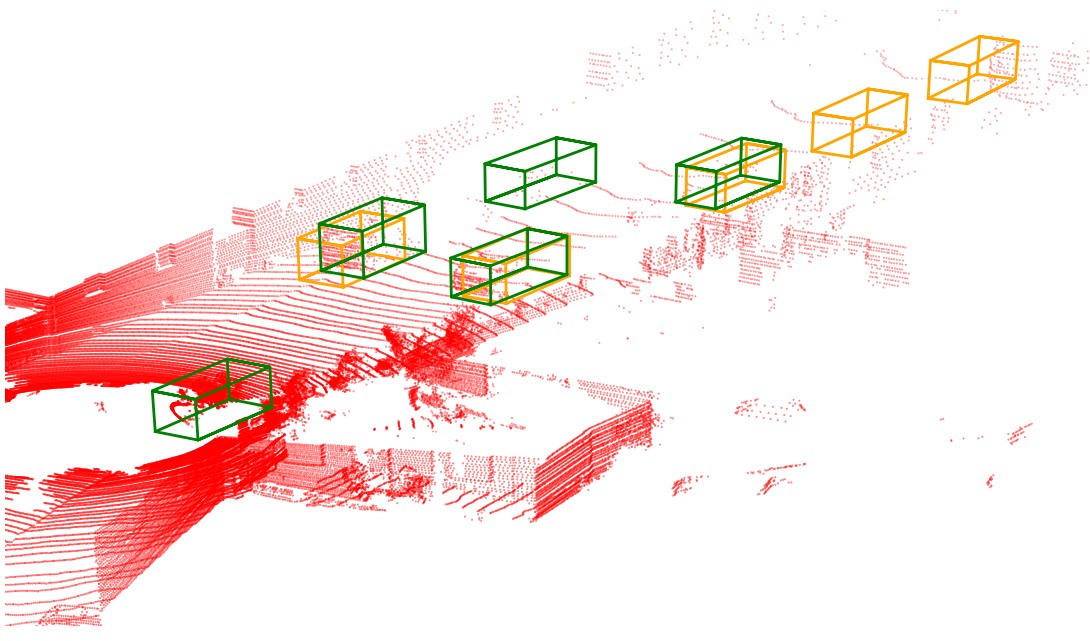}
\end{minipage}
\begin{minipage}{5.60cm}
   \includegraphics[width=1\linewidth]{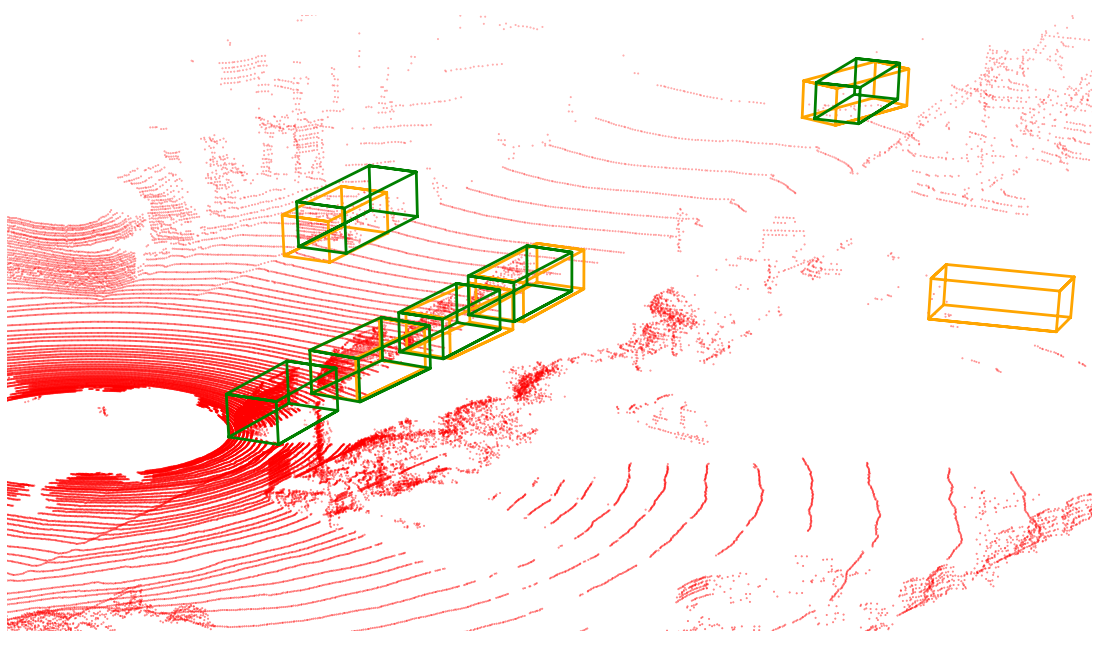}
\end{minipage}

\begin{minipage}{5.70cm}
   \includegraphics[width=1\linewidth]{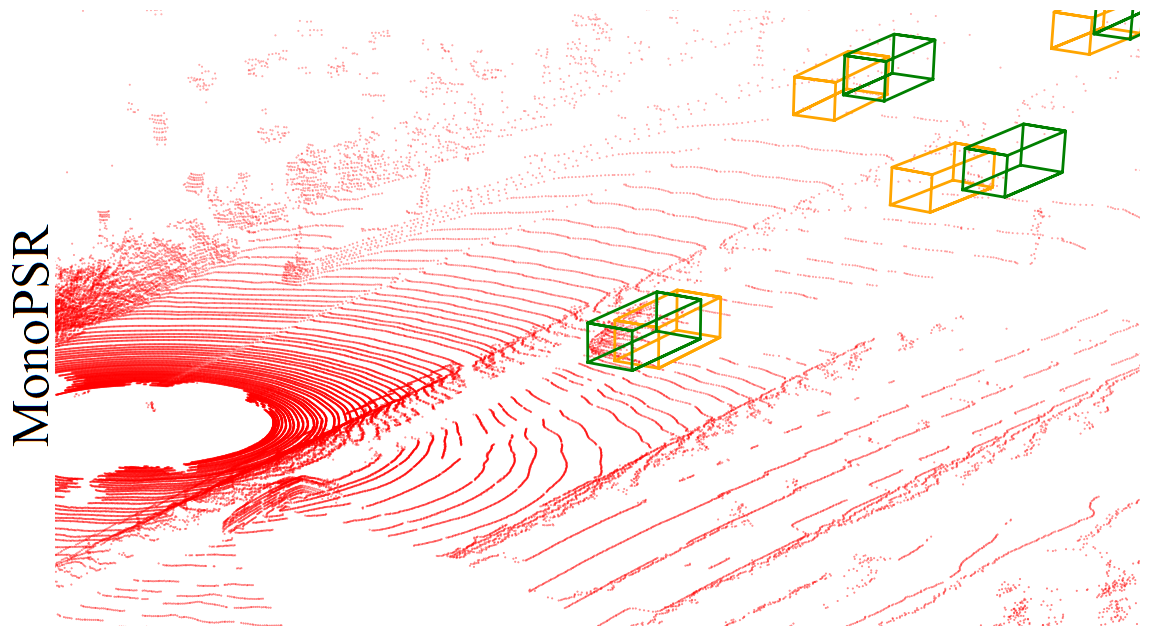}
\end{minipage}
\begin{minipage}{5.60cm}
   \includegraphics[width=1\linewidth]{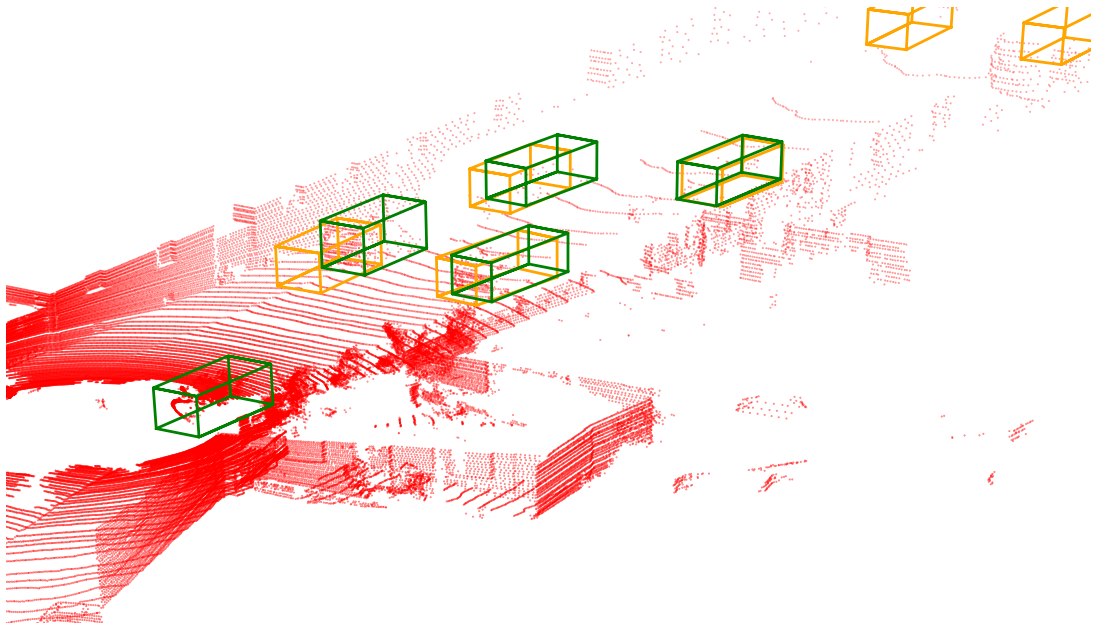}
\end{minipage}
\begin{minipage}{5.60cm}
   \includegraphics[width=1\linewidth]{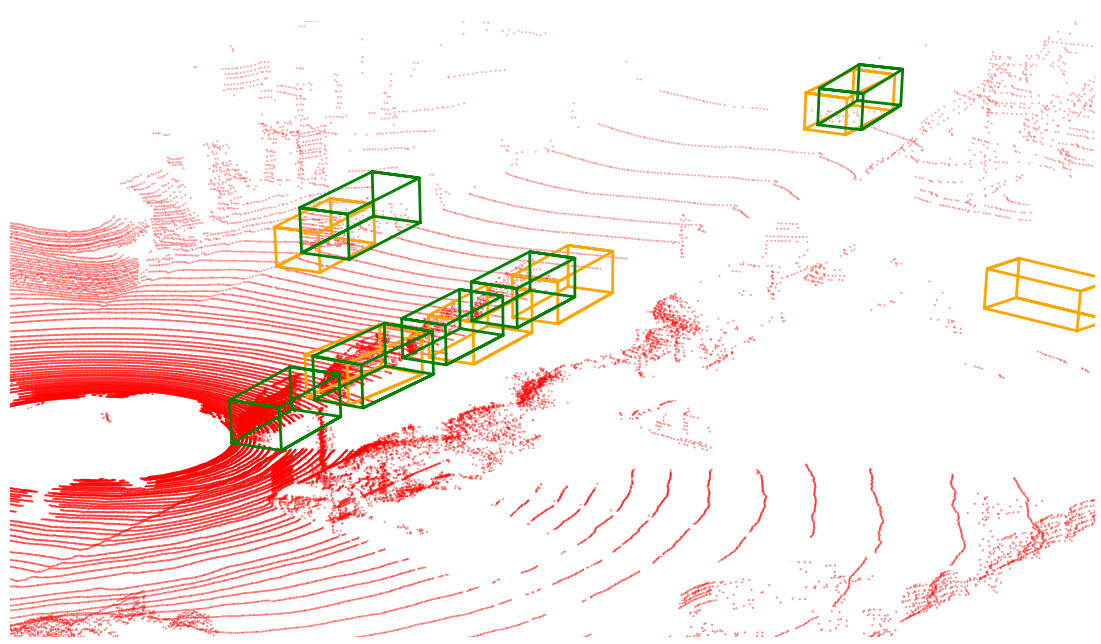}
\end{minipage}
\caption{The 3D detection results of MoNet3D for the different scenes in the KITTI benchmark dataset. For all the pictures, the green 3D boxes are the ground truth, the orange 3D boxes are the predictions, and the camera centres are in the bottom-left corner.}
\vskip -0.2in
\label{figure3}
\end{figure*}
\subsubsection{3D-Box Estimation}
This module of 3D box estimation predicts the centre point $C_{\mathrm{3d}} = (u^{(\mathrm{3d})}, v^{(\mathrm{3d})}, z^{(\mathrm{3d})})$ and vertices $\mathbf{\mathcal{O}}  = \{ \mathbf{O}_k\}_{k=1}^8$ of the 3D bounding box. To obtain the centre point $C_{\mathrm{3d}}$ of the 3D box, we inversely map the 2D box centre $C_{\rm{2d}}= (u^{(\mathrm{2d})}, v^{(\mathrm{2d})})$ through the camera calibration file provided by KITTI to obtain the coarse 3D position $C_{\mathrm{coa}}^{\mathrm{g}}$. The 2D to 3D inverse mapping expression is as follows:

\begin{equation}
\begin{cases}
u^{(\mathrm{3d})}=(u^{(\mathrm{2d})}-\theta)* \frac{\widehat{z}^\mathrm{g}}{f}\\
v^{(\mathrm{3d})}=(v^{(\mathrm{2d})}-\varphi)* \frac{\widehat{z}^\mathrm{g}}{f}
\end{cases}
\end{equation}
where $f$ is the focal length of the camera and $\theta$ and $\varphi$ are the main point parameters of the camera. For the coordinates of the 8 vertices of the 3D box $\mathbf{\mathcal{O}}=\{\mathbf{O}_k\}_{k=1}^8$, the method we choose is to directly use the deep features for regression. Similar to depth estimation, we also used shallow features to regress the offset $C_{\delta}^{\mathrm{g}}$ of the 3D  box centre points $C_{\rm{3d}}^{\mathrm{g}}$. The final $C_{\rm{3d}}^{\mathrm{g}}$ can be expressed as $C_{\rm{3d}}^{\mathrm{g}} = C _{\mathrm{coa}}^{\mathrm{g}} + C_{\delta}^{\mathrm{g}}$. The loss function for $C_{\rm{3d}}^{\mathrm{g}}$ and $\mathbf{\mathcal{O}}$ can be expressed as

\begin{align}
L_{\mathrm{3d}} &= \sum_\mathrm{g} \mathds{l}_\mathrm{g}^{\mathrm{obj}} \cdot \mathcal{L}_1 \left( \widehat{C}_{\mathrm{coa}}^\mathrm{g} + \widehat{C}_{\delta}^\mathrm{g} ,C_{\rm{3d}}^\mathrm{g} \right)+R(\mathrm{\mathbf{W}})&
 \\
L_{\mathcal{O}} &= \sum_\mathrm{g} \sum_k \mathds{l}_\mathrm{g}^{\mathrm{obj}} \cdot \mathcal{L}_1 \left( \widehat{\mathbf{O}}_k^\mathrm{g} , \mathbf{O}_k^\mathrm{g} \right)&
\end{align}
where $C_{\rm{3d}}^{\mathrm{g}}$ is the 3D centre point coordinate of the ground truth, $\widehat{C}_{\mathrm{coa}}^{\mathrm{g}}$ is the 3D centre point coordinate of the prediction from the deep features, $ \widehat{\mathbf{O}}_k^{\mathrm{g}}$ is the prediction of $\mathbf{O}_k^{\mathrm{g}}$, and $R(\mathrm{\mathbf{W}})$ refers to the regularization term that constrains the adjacent relationship of the object pair in 3D.

\section{Experiments}
We performed experiments on the KITTI dataset to verify and evaluate the effectiveness of our algorithm. Figure \ref{figure3} shows the visualization results of MoNet3D on the KITTI dataset. Pictures of three typical test scenarios are shown here, including high-speed roads, town roads, and neighbourhood roads. The pictures in lines 2 to 4 show the comparison between our proposed method and the latest object localization methods (MonoGNet \cite{qin2019monogrnet}, M3D \cite{xu2018multi}, and MonoPSR \cite{ku2019monocular}) and the 3D object detection results with the real detection results. In general, the MoNet3D method can effectively identify cars in 3D scenes, although in high-speed road scenes and town road scenes, some vehicle images have incomplete objects. Further observation reveals that M3D and MonoPSR have errors in long-distance object localization. In town road scenes, due to the consideration of the geometric similarity of adjacent objects, the MoNet3D method can better identify distant objects.
\subsection{Experiment Setup}

Most of the researches on 3D object detection of monocular cameras is verified on the KITTI dataset, so we also carry out experiments on the challenging dataset from KITTI to verify the effectiveness of the MoNet3D algorithm. We used the same method as Chen to split KITTI data sets into 3712 training images and 3769 testing images\cite{chen20153d}. The KITTI dataset contains three types of objects: easy (the bounding box height is greater than 40 pixels, all the objects are visible and truncated by no more than 15\%), moderate (the bounding box height is greater than 25 pixels, most objects are visible and truncated by no more than 30\%), hard (the bounding box height is greater than 25 pixels, and most of the objects are invisible and not truncated by more than 50\%).

Similar to other 3D object estimations, we used the localization and detection accuracy of automotive objects to verify the effectiveness of our method. In terms of object localization, the experiments calculated the relative accuracy of $u^{(\mathrm{3d})}, v^{(\mathrm{3d})}, z^{(\mathrm{3d})}$ as indicators; in terms of 3D object detection, the experiment used the average 3D accuracy rate and bird's-eye view average accuracy as indicators. For the car category, we compared the average accuracy of 3D object detection by the intersection over union (IOU) measure for different object types under two thresholds: 0.5 and 0.7.

We compared the experimental results of the proposed method on the KITTI dataset with state-of-the-art methods. The comparison methods include methods for extracting 3D object regional proposals, such as MF3D, ROI-10D, MonoPSR, and other latest methods for 3D object detection based on neural networks, such as Mono3D, Deep3Dbox, OFT-net, MF3D, ShiftNET, GS3D, and SS3D. We also compared MoNet3D with 3DOP, a 3D object detection method based on a binocular camera.

The experimental hyperparameter settings referred to MonGRNet. We initialized the model with random parameters. In the experiment, the similarity hyperparameter of Equation \ref{E1} was set to 100.00, and $\alpha$, $\beta$ and $\gamma$ were all set to 10.00. Model training uses tensorflow's SGD algorithm with momentum, batchsize is set to 2 and learning rate is set to $10^{-5}$. A total of 800000 iterations were trained on the KITTI dataset. Numerical experiments were performed on a computer equipped with an $\mathrm{Inter}^{\textregistered}$Core$\texttrademark$i7-6900K CPU, 32GB of memory, and an NVIDIA GeForce GTX 1080 Ti graphics card.
\subsection{Result of 3D Object Localization and Detection}
\begin{table*}[!htb]
\centering
\caption{\textbf{3D Detection}: Comparisons with the state-of-the-art methods in terms of the average precision for 3D object detection for the car category in the KITTI validation dataset with different IoUs}
\vskip 0.15in
\begin{center}
\setlength{\tabcolsep}{0.95mm}{
\begin{tabular}{c|ccc|ccc|c}
\hline
\multicolumn{1}{c|}{\multirow{2}{*}{Method}}     & \multicolumn{3}{c|}{AP3D (\textbf{IoU=0.5})} & \multicolumn{3}{c|}{AP3D (\textbf{IoU=0.7})}       & \multicolumn{1}{c}{\multirow{2}{*}{FPS$^a$}}         \\ \cline{2-7}
\multicolumn{1}{c|}{}                                                                   & Easy     & Moderate    & Hard     & Easy           & Moderate       & Hard    & \multicolumn{1}{c}{}       \\ \cline{1-8}
3DOP\cite{chen20173d}                                                                                                                     & 46.04    & 34.63       & 30.09    & 6.55           & 5.07           & 4.10       & 0.23    \\
Mono3D\cite{chen2016monocular}                                                                                                                      & 25.19    & 18.20       & 15.22    & 2.53           & 2.31           & 2.31         & 0.33   \\
OFT-Net\cite{roddick2018orthographic}                                                                                            & -                               & -                               & -                               & 4.07                            & 3.27             &3.29 &-                        \\
FQNet\cite{liu2019deep}                                                                                                & 28.16                               & 21.02                               & 19.91                               & 5.98                           & 5.50             &4.75 &2.00                        \\
ROI-10D\cite{manhardt2019roi}                                                                                      & -                               & -                               & -                               & 10.25                           & 6.39               &6.18        & -             \\
MF3D\cite{novak2017vehicle}                                                                                           & 47.88                           & 29.48                           & 26.44                           & 10.53                           & 5.69                            & 5.39                       & 8.33     \\
MonoDIS\cite{simonelli2019disentangling}                                                                                           & -                               & -                               & -                               & 11.06                            & 7.60             &6.37 &-                        \\
MonoPSR\cite{ku2019monocular}                                                                                      & 49.65                           & 41.71                           & 29.95                           & 12.75                           & 11.48                           & 8.59                   & 5.00           \\
ShiftNet\cite{naiden2019shift}                                                                                           & -                               & -                               & -                               & 13.84                           & 11.29                  &11.08         & -       \\
GS3D\cite{li2019gs3d}                                                                                             & 30.60                           & 26.40                           & 22.89                           & 11.63                           & 10.51                   &10.51  & 0.43            \\
SS3D\cite{jorgensen2019monocular}                                                                                            & -                               & -                               & -                               & 14.52                           & 13.15                           & 11.85                & 20.00           \\
M3D-RPN\cite{brazil2019m3d}                                                                                            & -                               & -                               & -                               & 20.27                            & 17.06             &15.21 &-                        \\
\textbf{Ours}                                                          & \textbf{55.64}$\pm$0.45    & \textbf{34.10}$\pm$0.14       & \textbf{34.10}$\pm$0.07    &\textbf{22.73}$\pm$0.30 & \textbf{16.73}$\pm$0.27 & \textbf{15.55}$\pm$0.24& \textbf{27.85}\\\hline
\end{tabular}}
\footnotesize{$^a$ FPS means frames per second and the FPS here refers to the FPS running on the computer.}
\end{center}
\vskip -0.1in
\label{table 1}
\end{table*}
In terms of object localization, we calculated the accuracy of the horizontal and height predictions estimated by the MoNet3D method and performed a comparison with the classic M3D. The experimental results showed that in the horizontal estimation direction ($u^{(\mathrm{3d})}$), M3D achieved a 90.59\% accuracy. Overall, in these three directions, our proposed method achieved an average accuracy of 96.07\%, where $u^{(\mathrm{3d})}$ is 94.74\%, $v^{(\mathrm{3d})}$ is 97.21\%, and $z^{(\mathrm{3d})}$ is 96.25\%. The experiments showed that because the horizontal regular optimization method was used, the proposed method was better than the recently proposed M3D in terms of the positioning accuracy of the horizontal direction.

\begin{figure}[t]
\vskip 0.2in
\begin{center}
   \includegraphics[width=1\linewidth]{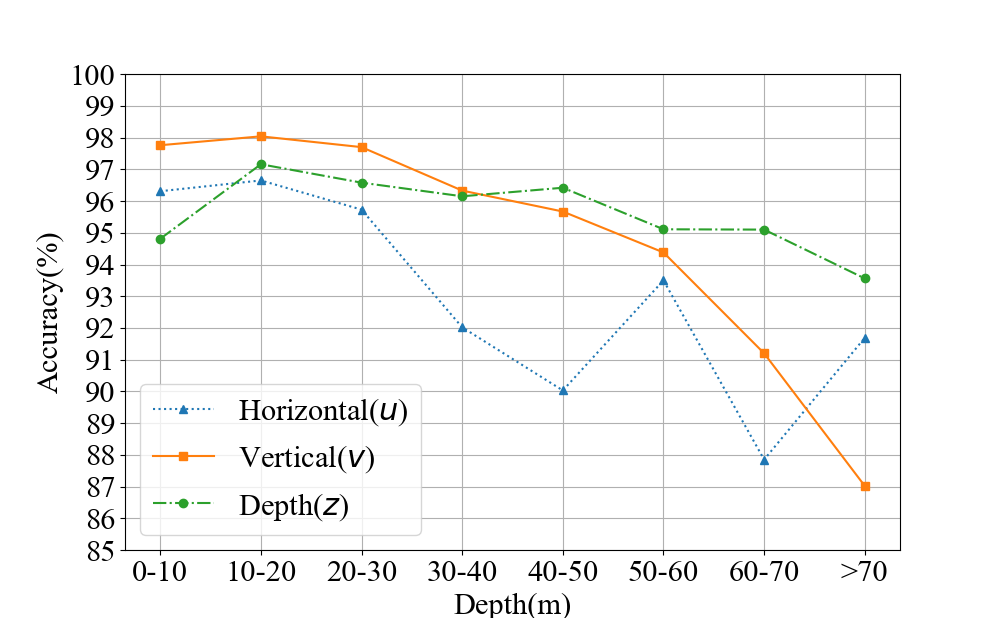}
\end{center}
\caption{Relative accuracy of the 3D box centre coordinates at different depths. The depth is divided into 5 groups of 10 metres, and the accuracy of $u^{(\mathrm{3d})},v^{(\mathrm{3d})},z^{(\mathrm{3d})}$ is calculated at different depths, where the blue line is the relative accuracy of the 3D box centre horizontal coordinate $u^{(\mathrm{3d})}$, the orange line is the relative accuracy of the 3D box centre vertical coordinate $v^{(\mathrm{3d})}$, and the green line is the relative accuracy of the 3D box centre depth coordinate $z^{(\mathrm{3d})}$.}
\vskip -0.2in
\label{figure4}
\end{figure}

Considering that most of the research on image depth estimation now was pixel-level depth estimation, we compared the instance-level depth estimation we invented with them. According to the latest research on pixel-level depth estimation\cite{fu2018deep,ren2019deep,liebel2019multidepth}, for example, the relative absolute error of the depth estimation of DORN on the KITTI dataset was 8.78\%\cite{fu2018deep}, and the relative absolute error of the depth estimation of MultiDepth on the same dataset was 13.82\%\cite{liebel2019multidepth}. Compared with pixel-level depth estimation method, the MoNet3D was coarser instance-level estimation method, and the depth estimation average accuracy was 96.25\%, which was significantly higher than the pixel-level depth estimation method.

To explore the effect of depth on the localization results, we grouped the test samples into groups of 10 metres (since the maximum distance of the car object in the KITTI dataset is 83 metres, and there are few objects with a depth of 80 metres or more, so we grouped 70-80 metres and 80-90 metres into one group) and evaluate the average accuracy of the MoNet3D method in $u^{(\mathrm{3d})}, v^{(\mathrm{3d})}, z^{(\mathrm{3d})}$. As shown in Figure \ref{figure4}, the average accuracy of the 3D box centre in the three directions is the largest when the depth is between 10 and 20 metres, and the accuracy in all directions decreases as the depth increases. However, even for objects 40 metres away, our proposed method still had a relative accuracy of more than 90\% in 3D object localization.

In addition to object localization, our proposed method can also perform 3D object detection, which is a very challenging task. Existing monocular 3D object detection methods have not achieved the accuracy of target recognition (see Table \ref{table 1}), but our research found that MoNet3D can still achieve better 3D recognition results under close-range conditions. When the IoU threshold was set to 0.3 and the depth was 0 to 10 metres and 10 to 20 metres, the accuracy of the proposed method in 3D object detection was 75.40\%-80.99\%. However, as the depth increased, as there was no other information, it was very difficult to predict the depth using only pictures whose depth information was severely compressed, and the prediction error also increased sharply. This experiment showed that our proposed method achieved good results in 3D object detection, but the current monocular 3D object detection method can only be applied to low-power ADASs (advanced driving-assistance systems) and other low-power embedded systems.
\section{Comparison with the State-of-the-Art Methods}
\begin{table*}[!htb]
\caption{\textbf{Bird's-Eye-View 3D Detection}: Comparisons with the state-of-the-art methods in terms of the 3D BEV(Bird's-Eye-View) and the inference time per image for the KITTI validation dataset with different IoUs.}
\vskip 0.15in
\begin{center}
\setlength{\tabcolsep}{1mm}{
\begin{tabular}{c|ccc|ccc|c}
\hline
\multicolumn{1}{c|}{\multirow{2}{*}{Method}}   & \multicolumn{3}{c|}{APBEV (\textbf{IoU=0.5})}               & \multicolumn{3}{c|}{APBEV (\textbf{IoU=0.7})}      & \multicolumn{1}{c}{\multirow{2}{*}{FPS$^a$}}         \\ \cline{2-7}
\multicolumn{1}{c|}{}                                                           & Easy           & Moderate       & Hard           & Easy           & Moderate       & Hard           & \multicolumn{1}{c}{} \\ \cline{1-8}
Mono3D\cite{chen2016monocular}                                                                                                              & 30.50          & 22.39          & 19.16          & 5.22           & 5.19           & 4.13      & 0.33       \\
FQNet\cite{liu2019deep}                                               & 32.57                               & 24.60                               & 21.25                               & 9.50                           & 8.02                            & 7.71                                 & -   \\
OFT-Net\cite{roddick2018orthographic}                                          & -                               & -                               & -                               & 11.06                           & 8.79                            & 8.91                                 & 2.00   \\
3DOP\cite{chen20173d}                                               & 55.04                               & 41.25                               & 34.55                               & 12.63                           & 9.49                            & 7.59                                 & 0.23   \\
ROI-10D\cite{manhardt2019roi}                                      & 46.9                            & 34.1                            & 30.5                            & 14.50                            & 9.9                             & 8.7                          & 5.00     \\
ShiftNet\cite{naiden2019shift}                                    & -                               & -                               & -                               & 18.61                           & 14.71                           & 13.57                               & -       \\
MonoPSR\cite{ku2019monocular}                                  & 56.97                           & 43.39                           & 36.00                           & 20.63                           & 18.67                           & 14.45                               & 5.00      \\
MF3D\cite{novak2017vehicle}                                          & -                               & -                               & -                               & 22.03                           & 13.63                           & 11.60                              & -     \\
\textbf{Ours}                                & \textbf{59.15}$\pm$0.20    & \textbf{43.26}$\pm$0.11       & \textbf{36.00}$\pm$0.06    &\textbf{27.48}$\pm$1.14 & \textbf{21.80}$\pm$0.29 & \textbf{17.86}$\pm$0.26& \textbf{27.85} \\ \hline
\end{tabular}}
\footnotesize{$^a$ FPS means frames per second and the FPS here refers to the FPS running on the computer.}
\end{center}
\vskip -0.1in
\label{table 2}
\end{table*}

To compare with other methods, we compared MoNet3D with recent monocular 3D object detection methods based on the KITTI dataset. The evaluation results are shown in Table \ref{table 1}. Overall, our proposed method reached the state-of-the-art level. It is clear from Table \ref{table 1} that our method significantly outperforms other methods in 3D object detection. When IoU = 0.3, the highest accuracy of our proposed method reached 72.56\%.

In addition to 3D object detection, we also evaluate the accuracy of our method in bird's-eye view (BEV) with IoU thresholds of 0.7 and 0.5 and compare it with recent monocular 3D object detection methods. The evaluation results are shown in Table \ref{table 2}. In general, the proposed method also surpasses other new methods recently proposed for BEV. Specifically, when IoU = 0.7, in easy mode, compared with other methods, the leading range is from 5.45\% to 22.26\%.

Embedded devices are often used in the field of autonomous driving, so there are very high requirements for energy efficiency. Due to the use of the highly efficient neural network, real-time image processing speed can be achieved. Compared with other methods, our regularization method does not bring speed loss with computational efficiency, which provides conditions for the application of embedded devices.
\section{System Performance for Embedded ADAS Applications}
To better explore the performance indicators of our method on embedded devices, such as the accuracy, real-time performance, and power, we apply the proposed method to an embedded NVIDIA Jetson AGX Xavier system. NVIDIA Jetson AGX Xavier mainly includes an 8-core NVIDIA Carmel ARMv8.2 64-bit CPU, a 512-core Volta architecture GPU consisting of 8 stream multiprocessors, 16 GB of memory, and an FP16 (computing power) of 11 TFLOPS (tera floating-point operations per second). Compared with the computer platform, Jetson AGX Xavier's storage capacity, computing power, and power consumption are far inferior, but our proposed method can be implemented in Jetson AGX Xavier with the same accuracy. The performance is over 5 frames per second(FPS), and the power at this time is 26.13 $W$.
\section{Discussion and Conclusion}
In this paper, we propose a novel monocular 3D object detection network. Using the proposed regularization term to optimize the corresponding loss function greatly enhances the ability of the original network in 3D object detection and pose estimation and improves the corresponding accuracy. It has excellent performance in 3D object detection, localization and attitude estimation. Moreover, the proposed method also has good migration ability, which can be applied to other networks to improve the accuracy of object detection of corresponding networks. In terms of the real-time performance, our method reaches 27.85 FPS, which is a fairly good real-time performance.

Of course, MoNet3D still has many limitations. First, object detection and localization is difficult using a monocular camera, so MoNet3D still has a series of limitations like other monocular camera object detection methods. Compared with methods based on radar and binocular cameras, 3D object detection methods using monocular cameras have lower accuracies. Current research finds that although 3D object detection methods for  monocular cameras have the advantage of low cost, they are only suitable for object localization and 3D object detection at short distances at low speeds. In high-speed scenarios, LiDAR must be combined to increase the accuracy to an acceptable level. Second, the current method is the same as ROI-1D, ShiftNet, MonoPSR, SS3D and other methods, and its accuracy is affected by 2D object detection. In the future, research on new methods of 2D object detection will improve the accuracy of 2D detection; on the other hand, using the information of 3D object detection to improve the accuracy of 2D object detection in turn will be considered.

\section*{Acknowledgements}
This work was supported by the National Natural Science Foundation of China under Contract 61971072.
\nocite{langley00}
\bibliography{example_paper}
\bibliographystyle{icml2020}
\end{document}